\documentclass[conference]{IEEEtran}
\IEEEoverridecommandlockouts
\usepackage{cite}
\usepackage{amsmath,amssymb,amsfonts}
\usepackage{algorithm}
\usepackage{algpseudocode}
\usepackage{graphicx}
\usepackage{textcomp}
\usepackage{xcolor}
\usepackage{booktabs}
\usepackage{multirow}
\usepackage{lipsum}
\usepackage{enumitem}
\usepackage{url}
\usepackage{hyperref}
\usepackage[hyphenbreaks]{breakurl}

\usepackage[T1]{fontenc}
\usepackage[utf8]{inputenc}

\def\BibTeX{{\rm B\kern-.05em{\sc i\kern-.025em b}\kern-.08em
    T\kern-.1667em\lower.7ex\hbox{E}\kern-.125emX}}
    
\begin{document}

\title{SalesRLAgent: A Reinforcement Learning Approach for Real-Time Sales Conversion Prediction and Optimization}

\author{\IEEEauthorblockN{Nandakishor M}
    \IEEEauthorblockA{
    Deepmost Innovations\\
    nandakishor@deepmostai.com
    }
}

\maketitle

\begin{abstract}
Current approaches to sales conversation analysis and conversion prediction typically rely on Large Language Models (LLMs) combined with basic retrieval augmented generation (RAG). These systems, while capable of answering questions, fail to accurately predict conversion probability or provide strategic guidance in real time. In this paper, we present SalesRLAgent, a novel framework leveraging specialized reinforcement learning to predict conversion probability throughout sales conversations. Unlike systems from Kapa.ai, Mendable, Inkeep, and others that primarily use off-the-shelf LLMs for content generation, our approach treats conversion prediction as a sequential decision problem, training on synthetic data generated using GPT-4O to develop a specialized probability estimation model. Our system incorporates Azure OpenAI embeddings, turn-by-turn state tracking, and meta-learning capabilities to understand its own knowledge boundaries. Evaluations demonstrate that SalesRLAgent achieves 96.7\% accuracy in conversion prediction, outperforming LLM-only approaches by 34.7\% while offering significantly faster inference. Furthermore, integration with existing sales platforms shows a 43.2\% increase in conversion rates when representatives utilize our system's real-time guidance. SalesRLAgent represents a fundamental shift from content generation to strategic sales intelligence, providing moment-by-moment conversion probability estimation with actionable insights for sales professionals.
\end{abstract}


\section{Introduction}
\label{sec:introduction}

The sales profession has increasingly embraced AI tools to enhance performance, with numerous platforms now offering chatbots, knowledge assistants, and conversation analytics. Despite these advances, existing solutions primarily focus on retrieving information or generating responses rather than providing strategic sales intelligence. Most commercial systems from providers like Kapa.ai, Mendable, and Inkeep function essentially as glorified RAG (Retrieval Augmented Generation) systems, connecting Large Language Models (LLMs) to company knowledge bases without truly understanding sales dynamics \cite{gao2023retrieval}.

These approaches face several fundamental limitations in sales contexts:

\begin{itemize}
    \item They cannot accurately predict conversion probability in real time
    \item They lack turn-by-turn tracking of how conversation dynamics affect likelihood of sale
    \item They provide generic rather than strategically-timed guidance
    \item They operate reactively to queries rather than proactively guiding sales strategy
    \item They have no meta-learning capability to understand the boundaries of their knowledge
\end{itemize}

In this paper, I present SalesRLAgent, a novel framework that reimagines sales AI as a specialized reinforcement learning system focused on conversion prediction and optimization. Rather than treating sales conversations as simple information retrieval problems, SalesRLAgent models them as sequential decision processes where each exchange impacts conversion probability.

The key contributions of this work include:

\begin{itemize}
    \item A reinforcement learning architecture specifically designed for sales conversation analysis and conversion prediction
    \item A synthetic data generation pipeline leveraging GPT-4O to create diverse and realistic sales conversations
    \item Novel state representation techniques using Azure OpenAI embeddings (3072 dimensions) with sales-specific features
    \item A meta-learning approach enabling the system to express confidence in its predictions based on conversation similarity to training data
    \item Integration mechanisms providing real-time guidance within existing sales platforms
    \item Extensive comparative evaluation demonstrating significant performance improvements over LLM-based approaches
\end{itemize}

Our findings show that specialized reinforcement learning models significantly outperform even the most advanced LLM-based systems in sales conversion tasks, with SalesRLAgent achieving 96.7\% prediction accuracy compared to 62\% for state-of-the-art LLM solutions. More importantly, when deployed in real-world sales environments, the system demonstrated a 43.2\% improvement in actual conversion rates.

\section{Related Work}
\label{sec:related_work}

\subsection{LLM-Based Sales Assistants}
Several commercial platforms have developed LLM-based sales assistants, including Kapa.ai \cite{kapa2023}, Mendable \cite{mendable2023}, and Inkeep \cite{inkeep2023}. These systems primarily utilize APIs from providers like OpenAI, Anthropic, and Google to connect their models with company knowledge bases. While effective for question answering, their architecture fundamentally limits their ability to model the complex, sequential nature of sales conversations \cite{zhang2023benchmarking}.

Gong's revenue intelligence platform \cite{gong2022} offers conversation analytics but relies primarily on pattern recognition rather than predictive modeling. Similarly, Chorus.ai \cite{chorus2021} provides post-conversation analysis but lacks real-time capabilities.

\subsection{Reinforcement Learning in Conversational AI}
Reinforcement Learning (RL) has shown promise in conversational tasks. Work by Li et al. \cite{li2020guided} demonstrated how RL can optimize dialogue policies, while Asghar et al. \cite{asghar2017deep} explored using RL for dialogue generation. However, these applications focused on general conversational quality rather than the specific dynamics of sales conversion.

In the sales domain, Takanobu et al. \cite{takanobu2019guided} proposed an RL framework for dialogue policy in task-oriented scenarios but did not address conversion prediction or real-time guidance. Henderson et al. \cite{henderson2018deep} highlighted the importance of proper experimental methodology in reinforcement learning applications, which guided our rigorous evaluation approach.

\subsection{Conversion Prediction Models}
Traditional conversion prediction models, as surveyed by Sakar et al. \cite{sakar2019real}, typically use classification approaches with features extracted from customer behavior data. The closest work to ours is from Yang et al. \cite{yang2022sequential}, who applied deep learning to optimize conversion funnels in e-commerce. However, their approach focused on web navigation rather than complex conversation dynamics, and lacked the meta-learning capabilities we've developed for uncertainty estimation.

To the best of my knowledge, no existing system has successfully combined reinforcement learning, Azure OpenAI embeddings, and meta-learning for real-time sales conversion prediction and guidance, making SalesRLAgent a novel contribution to both academic research and practical sales technology.

\section{Data Generation and Processing}
\label{sec:data}

\subsection{Dataset Construction}
Creating a high-quality dataset of sales conversations presented significant challenges. Unlike academic dialogue datasets, real sales conversations contain sensitive information and are rarely publicly available. To address this challenge, I developed a synthetic data generation approach:

\begin{itemize}
    \item Synthetic data generation using advanced prompting of GPT-4O
    \item Carefully designed templates for different sales scenarios across 15 industries
    \item Programmatic variation of conversation parameters (length, style, objection types)
    \item Controlled simulation of conversations using multiple LLM agents
\end{itemize}

Our final dataset comprised over 1.2 million synthetic conversations. I found that the quality of GPT-4O's output was sufficient to train effective models without requiring actual sales data. Each conversation included:

\begin{itemize}
    \item Complete conversation transcript
    \item Speaker information (customer vs. sales representative)
    \item Conversion outcome (binary)
    \item Timestamped conversion probability at each turn
    \item Simulated customer engagement metrics
    \item Product/service category and industry
\end{itemize}

Conversation lengths varied from 3 to 27 turns, with a median of 8 turns. The dataset had a roughly balanced distribution of positive (converted) and negative (non-converted) outcomes, with a slight bias toward negative outcomes (56\%) that reflects real-world sales dynamics.

\subsection{Data Processing and Embedding}
Raw conversation data required substantial preprocessing to create useful training examples for our reinforcement learning approach. Our pipeline included:

\begin{enumerate}
    \item Text cleaning and normalization
    \item Entity standardization across generated conversations
    \item Conversation segmentation into turns
    \item Feature extraction for each conversation turn
    \item Embedding generation using Azure OpenAI's embedding model
\end{enumerate}

For embeddings, we used Azure OpenAI's embedding model, which provides 3072-dimensional embeddings that effectively capture semantic relationships. Honestly, I initially wanted to build a custom embedding model, but after experimenting with the Azure OpenAI model, I found its performance was surprisingly good for our use case, so we stuck with it.

We processed both:
\begin{itemize}
    \item Semantic content (what was said)
    \item Conversational dynamics (how it was said)
\end{itemize}

This approach used the standard 3072-dimensional embeddings without customization, but with domain-specific feature engineering layered on top.

\subsection{State Representation Design}
A critical innovation in our approach is the state representation used for reinforcement learning. Rather than treating each conversation turn independently, we designed a state representation that captures the evolving dynamics of the conversation:

\begin{enumerate}
    \item Conversation history embeddings (weighted by recency and importance)
    \item Turn-specific features (speaking time, question density, sentiment)
    \item Customer engagement signals (response time, message length, question asking)
    \item Sales technique identification (SPIN selling, value selling, etc.)
    \item Objection and interest detection
\end{enumerate}

The complete state vector combined these elements into a comprehensive representation that enabled our RL agent to accurately model the sales process as a sequential decision problem.

\section{Reinforcement Learning Architecture}
\label{sec:architecture}

\subsection{Modeling Approach}
We formulate the sales conversion prediction task as a sequential decision problem where:

\begin{itemize}
    \item States represent the conversation at each turn
    \item Actions correspond to conversion probability estimates
    \item Rewards measure the accuracy of these predictions
\end{itemize}

This formulation allows us to leverage reinforcement learning to train a model that not only predicts final conversion outcomes but tracks conversion probability throughout the conversation.

I personally found that this sequential approach captured sales dynamics much more effectively than traditional classification models. Sales conversations often pivot on specific exchanges, and our model learned to identify these critical moments.

\subsection{Model Architecture}
The core of SalesRLAgent is a reinforcement learning architecture consisting of:

\begin{itemize}
    \item A state encoder network that processes Azure OpenAI embeddings and features
    \item A policy network that estimates conversion probability based on the current state
    \item A value network that estimates the expected cumulative reward
    \item A meta-learning module that assesses prediction confidence
\end{itemize}

We experimented with several RL algorithms, ultimately finding that a specialized reinforcement learning algorithm with the following characteristics performed best:

\begin{enumerate}
    \item Strong policy regularization to prevent overfitting
    \item Conservative policy updates to maintain stability
    \item Adaptive learning rates based on prediction difficulty
    \item Distribution-aware learning to handle uncertainty
\end{enumerate}

The model architecture includes several specialized components:

\begin{algorithm}
\caption{Conversion Probability Estimation}
\begin{algorithmic}[1]
\Function{EstimateConversion}{$conversation$, $turn$}
    \State $history \gets$ ExtractHistory($conversation$, $turn$)
    \State $embeddings \gets$ GenerateAzureEmbeddings($history$)
    \State $features \gets$ ExtractFeatures($history$)
    \State $state \gets$ CombineState($embeddings$, $features$)
    \State $policy \gets$ PolicyNetwork($state$)
    \State $value \gets$ ValueNetwork($state$)
    \State $confidence \gets$ ConfidenceEstimation($state$, $policy$)
    \State \Return $\{probability: policy, confidence: confidence\}$
\EndFunction
\end{algorithmic}
\end{algorithm}

\subsection{Meta-Learning for Uncertainty Estimation}
A key innovation in SalesRLAgent is its meta-learning capability \cite{finn2017model}. Unlike typical black-box models, our system can estimate its own confidence in predictions based on:

\begin{itemize}
    \item Similarity to conversations in the training data
    \item Consistency of predictions across model ensembles
    \item Pattern recognition of conversation structures
    \item Identification of novel elements not present in training
\end{itemize}

This meta-learning component allows SalesRLAgent to know when it doesn't know—a critical capability for production systems. When the system encounters conversations with unfamiliar patterns, it can explicitly communicate lower confidence rather than making potentially misleading predictions.

I spent countless late nights tuning this meta-learning approach, motivated by early deployments where I noticed the model sometimes made confident but incorrect predictions on unusual conversation patterns. The resulting uncertainty estimation has proven invaluable in production systems.

\subsection{Training Methodology}
Training SalesRLAgent required a carefully designed approach to handle the diverse and sometimes noisy nature of synthetically generated conversation data. Our training procedure included:

\begin{enumerate}
    \item Initial supervised learning phase to establish reasonable policy initialization
    \item Reinforcement learning with carefully constructed reward functions
    \item Curriculum learning, starting with simpler conversations
    \item Adversarial training with challenging counter-examples
    \item Ensemble techniques to improve robustness
    \item Specialized batch construction to balance conversation types
\end{enumerate}

We trained on a standard CPU infrastructure, which honestly surprised me by how well it performed. The entire training process took approximately 6 hours, which was much faster than I initially expected given the complexity of the model and the size of our dataset.

\section{System Integration and Deployment}
\label{sec:integration}

\subsection{Real-time Inference Pipeline}
Converting our trained model into a production system required careful engineering. The complete inference pipeline includes:

\begin{enumerate}
    \item Conversation capturing and preprocessing
    \item Incremental embedding generation using Azure OpenAI
    \item State tracking and update
    \item Probability prediction and confidence estimation
    \item Strategic guidance generation
    \item CRM and communication tool integration
\end{enumerate}

To meet stringent latency requirements (under 100ms), we implemented several optimizations:

\begin{itemize}
    \item Model quantization (8-bit precision with minimal accuracy loss)
    \item Embedding caching for repeated conversation elements
    \item Incremental state updates rather than full recalculation
    \item CPU optimization for multi-user environments
    \item Asynchronous guidance generation
\end{itemize}

\subsection{Advanced Orchestration Layer}
A critical component of our system is a sophisticated orchestration layer that handles multi-node context retrieval and processing. This directed graph architecture routes conversations through specialized processing nodes for contextually relevant information retrieval.

Our implementation integrates:
\begin{itemize}
    \item High-performance vector search for similarity matching
    \item Caching layers for optimized retrieval
    \item State-based workflow orchestration for complex conversation patterns
    \item Dynamic prompt engineering based on conversion probability
\end{itemize}

The orchestration architecture enables the system to make sophisticated decisions about what information to retrieve and how to process it, significantly outperforming simple RAG approaches used by competing systems.

\subsection{CRM and Communication Platform Integration}
SalesRLAgent was designed to integrate with existing sales tooling rather than requiring wholesale replacement of infrastructure. We developed connectors for:

\begin{itemize}
    \item Popular CRM platforms (Salesforce, HubSpot, etc.)
    \item Communication tools (Zoom, Teams, Google Meet)
    \item Email platforms (Outlook, Gmail)
    \item Chat systems (Slack, Intercom, Drift)
\end{itemize}

Integration options include:
\begin{itemize}
    \item Real-time overlay for sales representatives
    \item Post-call summary and analytics
    \item Manager dashboards for team performance
    \item API access for custom integrations
\end{itemize}

\section{Results and Evaluation}
\label{sec:results}

\subsection{Conversion Prediction Accuracy}

We evaluated SalesRLAgent against several baseline approaches:

\begin{itemize}
    \item Traditional ML classifiers (Random Forest, XGBoost)
    \item LLM-based prediction (GPT-4 with specialized prompting)
    \item Commercial systems (anonymized due to licensing restrictions)
    \item Hybrid approaches combining LLMs with traditional ML
\end{itemize}

Table \ref{tab:accuracy} shows the prediction accuracy across these approaches:

\begin{table}[!t]
\caption{Conversion Prediction Accuracy}
\label{tab:accuracy}
\centering
\begin{tabular}{lcc}
\toprule
\textbf{Approach} & \textbf{Accuracy} & \textbf{AUC-ROC} \\
\midrule
Random Forest & 0.67 & 0.71 \\
XGBoost & 0.69 & 0.74 \\
GPT-4 (zero-shot) & 0.59 & 0.63 \\
GPT-4 (few-shot) & 0.62 & 0.68 \\
Commercial System A & 0.71 & 0.76 \\
Commercial System B & 0.73 & 0.78 \\
Hybrid LLM+ML & 0.75 & 0.81 \\
\midrule
SalesRLAgent (base) & 0.92 & 0.94 \\
SalesRLAgent (full) & \textbf{0.967} & \textbf{0.98} \\
\bottomrule
\end{tabular}
\end{table}

SalesRLAgent achieved 96.7\% accuracy, outperforming the best commercial alternative by 23.7 percentage points and the best LLM approach by 34.7 percentage points.

More importantly, SalesRLAgent demonstrated superior performance in tracking conversion probability throughout conversations, not just predicting final outcomes.

\subsection{Real-World Impact on Sales Performance}

Beyond technical metrics, we evaluated SalesRLAgent in real-world sales environments through A/B testing. Sales representatives were randomly assigned to use either:

\begin{itemize}
    \item Traditional sales tools only (control group)
    \item Traditional tools + SalesRLAgent guidance (test group)
\end{itemize}

After 90 days across 217 representatives and 12,433 conversations, we observed:

\begin{itemize}
    \item 43.2\% increase in conversion rate for the test group
    \item 22\% reduction in sales cycle length
    \item 14\% improvement in average deal size
    \item 9\% increase in customer satisfaction scores
\end{itemize}

Qualitative feedback from sales representatives highlighted several key benefits:

\begin{enumerate}
    \item Early identification of promising leads, allowing better time allocation
    \item Real-time awareness of conversation turning points
    \item Specific guidance on addressing objections
    \item Increased confidence in forecasting
    \item Better alignment between perception and reality of conversation quality
\end{enumerate}

\subsection{Inference Performance}

For practical sales tools, inference speed is critical. We benchmarked SalesRLAgent against LLM-based approaches:

\begin{table}[!t]
\caption{Inference Performance Comparison}
\label{tab:inference}
\centering
\begin{tabular}{lcc}
\toprule
\textbf{Approach} & \textbf{Latency (ms)} & \textbf{Throughput (req/s)} \\
\midrule
GPT-4 (API) & 3450 & 0.3 \\
GPT-3.5 (API) & 980 & 1.0 \\
Claude (API) & 2750 & 0.4 \\
Local LLM & 1650 & 0.6 \\
\midrule
SalesRLAgent (CPU) & \textbf{85} & \textbf{12} \\
\bottomrule
\end{tabular}
\end{table}

SalesRLAgent achieved dramatically faster inference times—85ms compared to seconds for LLM approaches—making it suitable for real-time guidance without disrupting conversation flow. We currently only offer CPU deployment options, which has proven more than sufficient for our throughput needs.

\subsection{Ablation Studies}

To understand the contribution of each component, we conducted ablation studies removing key elements of SalesRLAgent:

\begin{table}[!t]
\caption{Ablation Study Results (Accuracy)}
\label{tab:ablation}
\centering
\begin{tabular}{lc}
\toprule
\textbf{Configuration} & \textbf{Accuracy} \\
\midrule
Full SalesRLAgent & 0.967 \\
- Azure OpenAI embeddings & 0.89 (-0.077) \\
- Sequential modeling & 0.86 (-0.107) \\
- Meta-learning & 0.94 (-0.027) \\
- Advanced orchestration & 0.92 (-0.047) \\
- Caching optimization & 0.95 (-0.017) \\
- All specializations (basic ML) & 0.68 (-0.287) \\
\bottomrule
\end{tabular}
\end{table}

These results highlight the significant contribution of sequential modeling and Azure OpenAI embeddings, which together account for the majority of SalesRLAgent's performance advantage over traditional approaches. The advanced orchestration and caching optimization also contribute substantially to overall system performance.

\section{Discussion and Insights}
\label{sec:discussion}

\subsection{Comparative Analysis with Existing Solutions}

Through extensive testing, I've identified several key advantages of our reinforcement learning approach compared to LLM-based systems like those from Kapa.ai, Mendable, and Inkeep:

\begin{enumerate}
    \item \textbf{Specialized vs. Generic}: While LLMs excel at general text generation, they lack the specialized training on sales dynamics that our RL approach provides. This specialization results in significantly higher prediction accuracy and more relevant guidance.
    
    \item \textbf{Sequential vs. Static}: LLM-based systems typically treat each interaction independently, missing the crucial sequential dynamics of sales conversations. Our RL framework explicitly models how conversations evolve over time.
    
    \item \textbf{Computationally Efficient vs. Resource-Intensive}: LLM inference requires significant computational resources and incurs high latency. SalesRLAgent delivers predictions in milliseconds rather than seconds, enabling truly real-time guidance.
    
    \item \textbf{Quantitative vs. Qualitative}: LLM systems typically provide qualitative guidance without concrete probability estimates. SalesRLAgent offers precise conversion probabilities with confidence intervals, enabling more data-driven sales strategies.
    
    \item \textbf{Meta-Learning vs. Black Box}: Our system's meta-learning capability allows it to express uncertainty when appropriate, unlike LLM systems that often present speculation as fact.
\end{enumerate}

These differences reflect fundamentally different approaches to sales AI. While LLM-based systems function primarily as information retrieval and text generation tools, SalesRLAgent operates more like a chess engine for sales—analyzing conversation position, evaluating conversion probability, and recommending optimal moves.

\subsection{Limitations and Future Work}

Despite its strong performance, SalesRLAgent has several limitations that represent opportunities for future work:

\begin{itemize}
    \item \textbf{Multilingual Support}: The current model primarily supports English, with only limited capabilities in other languages. Expanding to a truly multilingual model presents unique challenges in both data generation and modeling.
    
    \item \textbf{Multimodal Analysis}: The system currently analyzes only text, missing important signals from voice tone, facial expressions, or gestures in video calls. Incorporating these signals represents a promising direction for improvement.
    
    \item \textbf{Personalization}: While the model adapts to conversation dynamics, it does not yet personalize to individual sales representatives' styles and strengths. Adding this layer of personalization could further enhance effectiveness.
    
    \item \textbf{Causality vs. Correlation}: The current model identifies correlations between conversation patterns and outcomes but has limited capability to determine causality. Strengthening causal reasoning represents an important frontier.
    
    \item \textbf{Long-term Relationship Modeling}: The system currently focuses on individual conversations rather than modeling customer relationships over multiple interactions. Extending to relationship-level prediction presents interesting challenges.
\end{itemize}

I'm particularly excited about the potential for multimodal analysis, as my preliminary experiments suggest that incorporating audio features could improve accuracy by an additional 2-3\% beyond our current 96.7\% benchmark.

\section{Conclusion}
\label{sec:conclusion}

This paper presented SalesRLAgent, a novel reinforcement learning approach to sales conversion prediction and optimization. Unlike existing LLM-based systems that rely on general text generation capabilities, SalesRLAgent treats sales conversations as sequential decision processes, enabling accurate conversion probability tracking and strategic guidance.

The key innovations in our approach include:
\begin{itemize}
    \item A reinforcement learning framework specifically designed for sales conversion dynamics
    \item Synthetic data generation using GPT-4O to create diverse training scenarios
    \item Azure OpenAI 3072-dimensional embeddings for conversation representation
    \item Advanced orchestration for complex conversation flow management
    \item Optimized vector similarity search with caching for millisecond-level retrieval
    \item Meta-learning capabilities for confidence estimation and uncertainty quantification
    \item Real-time integration with existing sales tools and platforms
\end{itemize}

Our evaluation demonstrated that SalesRLAgent achieves 96.7\% accuracy in conversion prediction, significantly outperforming both traditional machine learning approaches and LLM-based alternatives. More importantly, when deployed in real-world sales environments, the system drove a 43.2\% increase in conversion rates and a 22\% reduction in sales cycle length.

These results suggest that specialized reinforcement learning approaches have significant advantages over generic LLM solutions for domain-specific applications like sales. While large language models excel at general text generation, the complex, sequential nature of sales conversations benefits from the targeted optimization that reinforcement learning provides.

Looking ahead, I believe this work opens new possibilities for AI in sales—moving beyond simple automation and information retrieval toward strategic partnership that enhances human capabilities rather than replacing them. The future of sales AI lies not in increasingly large language models, but in increasingly specialized intelligence that deeply understands the dynamics of effective selling.

\section*{Acknowledgment}
I would like to express my sincere gratitude to the team at Deepmost Innovations who endured countless iterations of model design and my incessant whiteboard sessions about reinforcement learning architecture. Special thanks to the anonymous reviewers whose constructive feedback significantly improved this paper, though any remaining errors or oversights are entirely my own. I'd also like to acknowledge OpenAI for providing the GPT-4O model that made our synthetic data generation possible, and Microsoft for the Azure OpenAI embedding model that formed the foundation of our system's conversational understanding. Finally, I must acknowledge the thousands of cups of coffee that fueled late-night debugging sessions and the patient family members who supported this work, even when they couldn't quite understand my excited ramblings about conversion probability distributions at the dinner table.

\end{document}